\documentclass[letterpaper]{article} 
\usepackage{aaai23}  
\usepackage{times}  
\usepackage{helvet}  
\usepackage{courier}  
\usepackage[hyphens]{url}  
\usepackage{graphicx} 
\usepackage{natbib}
\usepackage{caption} 
\usepackage{algorithm}
\usepackage{algorithmic}
\usepackage{bm}
\newcommand{\kl}[2]{D_{\mathrm{KL}}\left[\,#1\,\|\,#2\,\right]}
\usepackage[normalem]{ulem}
\useunder{\uline}{\ul}{}
\usepackage{amsmath}
\usepackage{newfloat}
\usepackage{listings}
\DeclareCaptionStyle{ruled}{labelfont=normalfont,labelsep=colon,strut=off} 
\floatstyle{ruled}
\newfloat{listing}{tb}{lst}{}
\floatname{listing}{Listing}

\title{AltUB: Alternating Training Method to Update Base Distribution of \\Normalizing Flow for Anomaly Detection}
\author{
    Yeongmin Kim\equalcontrib \textsuperscript{\rm 1}, 
    Huiwon Jang\equalcontrib \textsuperscript{\rm 2}, 
    DongKeon Lee \textsuperscript{\rm 3}, 
    Ho-Jin Choi \textsuperscript{\rm 3}
}
\affiliations {
    \textsuperscript{\rm 1} School of Freshman, KAIST\\
    \textsuperscript{\rm 2} Department of Mathematical Science, KAIST\\
    \textsuperscript{\rm 3} School of Computing, KAIST\\
    (cytotoxicity8 / huiwoen0516 / hagg30 / hojinc)@kaist.ac.kr
}

\begin{document}

\maketitle

\begin{abstract}
Unsupervised anomaly detection is coming into the spotlight these days in various practical domains due to the limited amount of anomaly data. One of the major approaches for it is a normalizing flow which pursues the invertible transformation of a complex distribution as images into an easy distribution as $N(0, I)$. In fact, algorithms based on normalizing flow like FastFlow and CFLOW-AD establish state-of-the-art performance on unsupervised anomaly detection tasks. Nevertheless, we investigate these algorithms convert normal images into not $N(0, I)$ as their destination, but an arbitrary normal distribution. Moreover, their performances are often unstable, which is highly critical for unsupervised tasks because data for validation are not provided. To break through these observations, we propose a simple solution AltUB which introduces alternating training to update the base distribution of normalizing flow for anomaly detection. AltUB effectively improves the stability of performance of normalizing flow. Furthermore, our method achieves the new state-of-the-art performance of the anomaly segmentation task on the MVTec AD dataset with 98.8\% AUROC.
\end{abstract}

\section{Introduction}
Automated detection of anomalies has become an important area of computer vision in a variety of practical domains, including industrial \cite{bergmann1}, medical \cite{zhou1} field, and autonomous driving systems \cite{jung1}. The essential goal of anomaly detection (AD) is to classify whether an object is normal or abnormal, and localize the regions of an anomaly if the object is classified as an anomaly \cite{wu1}. However, it is challenging to obtain a large number of abnormal images in real-world problems.

An unsupervised anomaly detection task has been introduced to address this imbalance in the dataset due to the lack of defected samples. To do so, it aims at detecting defection by using only the non-defected samples. That is why many recent approaches for unsupervised anomaly detection try to obtain representations of normal data and detect anomalies by comparing them with test samples' representations.

Learning representations of normal data in an unsupervised anomaly detection task is closely related to the goal of normalizing flow: the model learns invertible mapping of a complex distribution of normal samples into a simple distribution such as the normal distribution. The original usages of the normalizing flow are density estimation and generating data efficiently using invertible mappings \cite{dinh1, kingma1}, whereas some recent approaches \cite{rudolph1, yu1, gudovskiy1} begin to utilize the normalizing flow to tackle the unsupervised anomaly detection tasks. In particular, they have achieved state-of-the-art performance on the MVTec AD \cite{bergmann1} of the industrial domain.

\begin{figure*}[tb]
\begin{center}
    \includegraphics[width=\textwidth]{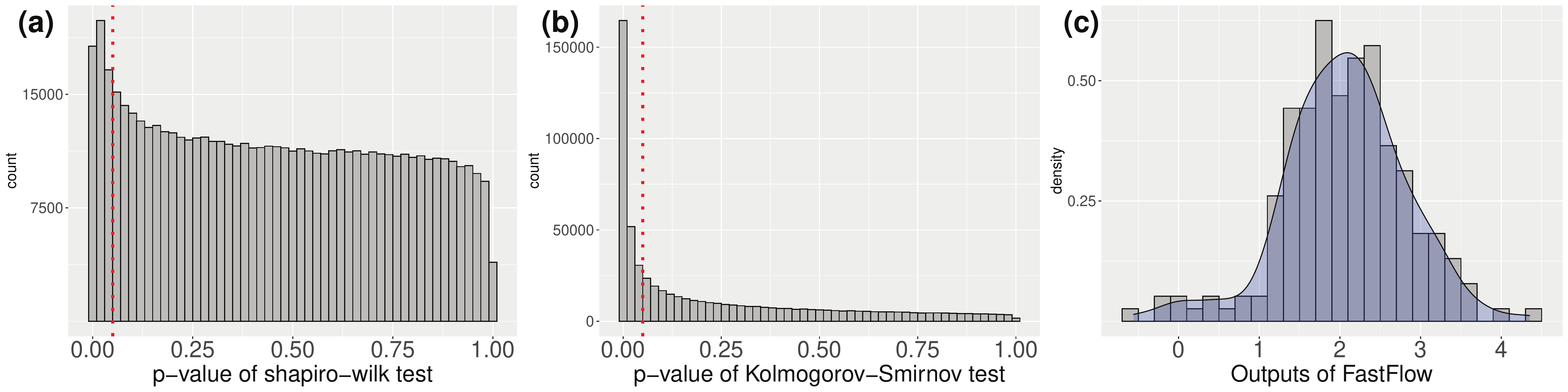}
\caption{Histograms of outputs of FastFlow. This figure shows that outputs of normalizing flow follow an arbitrary normal distribution, not the base distribution $N(0, I)$.
Feature extractor: CaiT, input: MVTec AD capsule category.
(a) The result of the Shapiro-Wilk test \cite{shapiro} applied to values of each spatial location of each channel among outputs from normal images. The distribution with a p-value $> 0.05$ is usually considered normal.
(b) The result of the Kolmogorov-Smirnov test (KS-test) \cite{kstest} applied to the same ones with (a). p-value $>0.05$ implies the distribution is highly likely to be $N(0, 1)$.
(c) Visualization of one of the distributions. Ideally, it should follow $N(0, 1)$. However its distribution is roughly $N(2, 0.5)$.}
\label{figure-fastflow-dist}
\end{center}
\end{figure*}

Despite the success of normalizing flow-based anomaly detection models, they have a limitation for the unsupervised task: performances are unstable in many cases. Specifically, their test performances often fluctuate while training the model for a long time (e.g. 200 or 400 epochs). This drawback will be critical in every practical domain because one must use the unsupervised trained models that have been trained sufficiently. One possible reason is the limited expressive power of normalizing flow with the fixed base (prior) distribution. Current approaches as \citet{rudolph1, yu1, gudovskiy1} fix the base distribution of normalizing flow as $N(0, I)$. However, as depicted in Figures \ref{figure-fastflow-dist} (b) and (c), most outputs of the model do not follow $N(0, I)$, while they utilize the likelihoods in the base distribution as a score function: 
\begin{equation}
    -\exp\left({\frac{1}{|C|}}\sum_{c \in C}{-\frac{z_c^Tz_c}{2}}\right),
\end{equation}
where $C$ is an index set for output channels. Even the score function requires an assumption for the theoretical validity that normal samples are transformed to the base distribution while training, it fails for the models to learn $N(0, I)$. This phenomenon occurs pervasively for normalizing flow-based anomaly detection models.

\begin{figure}[t]
\centering
\includegraphics[width=\columnwidth]{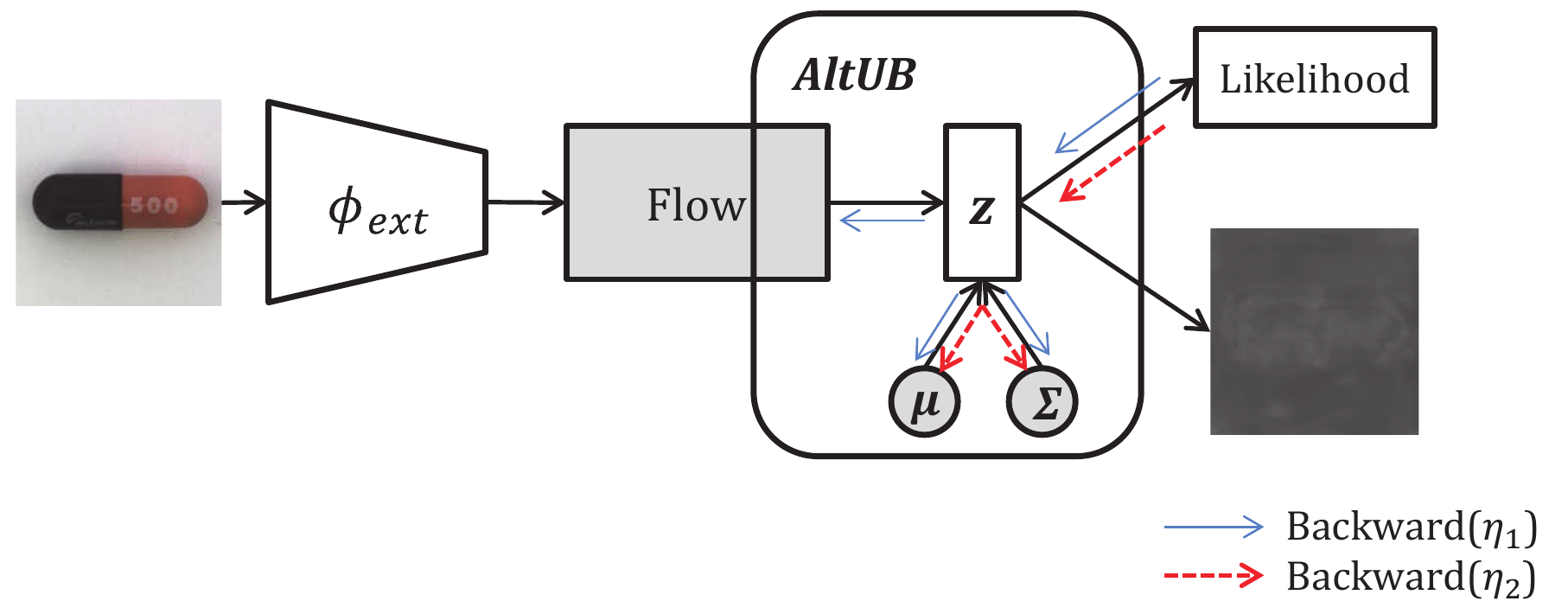}

\caption{The training process of AD flow+AltUB. The shaded (Flow, $\mu, \Sigma$) are trainable, and the two backward lines (blue solid line and red dotted line) indicate the alternating updates. While training, the model learns so that likelihood is high and the anomaly score is low for normal samples.}
\label{fig:2}
\end{figure}

\begin{figure*}[t]
\begin{center}
    \includegraphics[width=\textwidth]{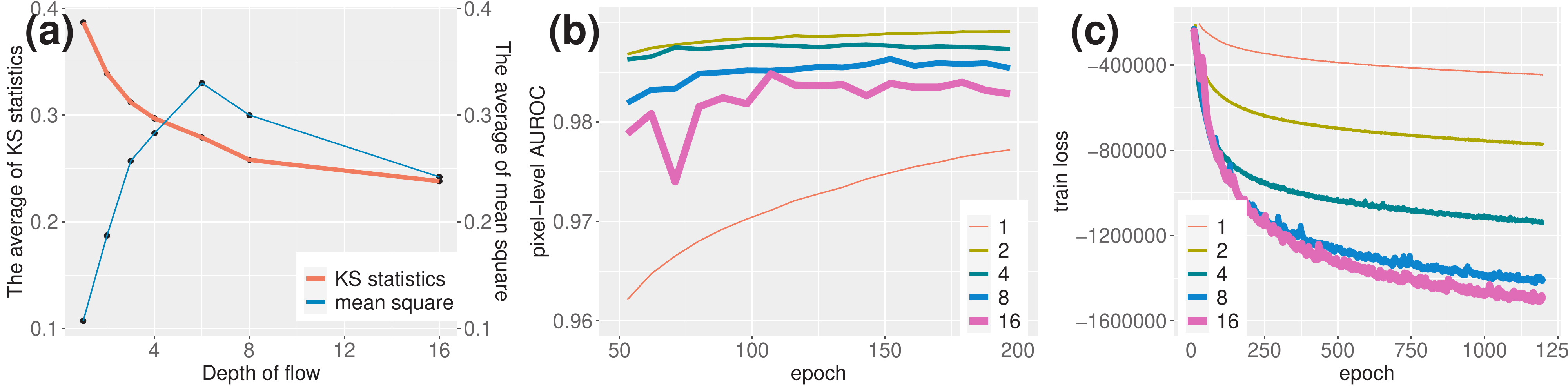}
\caption{Various statistics and performance depend on the depth of normalizing flow. FastFlow with R18 is applied to the capsule category on MVTec AD.
(a) The average of Kolmogorov-Smirnov statistics and mean square of outputs at 200th epochs. These statistics are calculated from each CHW dimension.
(b) The pixel-level AUROCs were measured during the training. ($1 < 16 < 8 < 4 < 2$)
(c) The training loss measured during the training. ($16 < 8 < 4 < 2 < 1$)
}
\label{figure-fastflow-depth}
\end{center}
\end{figure*}

To tackle this problem by considering the reason mentioned above, we propose an algorithm called the \textit{\textbf{Alt}ernating Training method to \textbf{U}pdate \textbf{B}ase distribution (AltUB)} that performs better and learns more stably with normalizing flow-based anomaly detection models. Our method introduces alternating updates to learn a base distribution more actively and to allow the model to adapt to the changed base distribution gradually. This simple but effective updating scheme improves the expressive power of the normalizing flow, especially for anomaly detection.

To verify the stability of our method under the unsupervised anomaly detection task, we measure the average performance for some training epoch intervals as well as the best performance. In the experimental results, our method improves the stability of normalizing flow-based anomaly detection models. In addition, our method achieves state-of-the-art performance on anomaly segmentation of both MVTec AD and BTAD datasets. These results support that our method is learning-stable and performing better.

\subsubsection{Contributions} In summary, we propose AltUB for the anomaly detection tasks with the following contributions.
\begin{itemize}
    \item We investigate that normalizing flow models for anomaly detection commonly fail to transform normal images into $N(0, I)$.
    \item We suggest the update of the base distribution can solve the above defect, and propose a proper method to update the base distribution: Alternating training.
    \item Our model achieves state-of-the-art performance on anomaly detection of MVTec AD and BTAD datasets.
\end{itemize}

\section{Related work}
\subsection{Normalizing flow-based anomaly detection models}
Unsupervised anomaly detection models based on normalizing flow have shown high performance on various practical tasks \cite{rudolph1, yu1, gudovskiy1}. Normalizing flow-based models first obtain representations of normal images from a pre-trained feature extractor. They utilize the representations to learn a distribution of normal data. Because of their simple yet powerful idea of invertible mappings, they effectively describe the distribution. To detect anomalies, they estimate the likelihood of the base distribution of test samples.

However, their expressive power is not enough as shown in Figure \ref{figure-fastflow-dist}. It might cause of their low stability on performance while learning i.e. performance decreases as learning progresses. In this paper, we suggest a simple method to increase the expressive power of normal data to improve the AUROC score.

\subsection{Methods to update the base distribution of normalizing flow}
Some prior works \cite{kingma1, mar-scf} have suggested methods to train the base distribution of normalizing flow for image generation and density estimation. Glow \cite{kingma1}, which is the well-known model for image generation, tries to update the channel-wise normal base distribution through a single layer of convolution neural network. Also, mAR-SCF \cite{mar-scf} applies the multi-scale autoregressive prior by using ConvLSTMs to increase the expressive power of split coupling layers.

However, these approaches are not appropriate for anomaly detection tasks for the following reasons. First, the output seems to follow not only channel-wise but also spatial dimensional normal distribution. Second, the layers can't distinguish normal images and abnormal ones because these layers train not parameters of distribution but the way to get them from inputs. Note that reflection of the sample statistics from anomaly disrupts the detection.

Nevertheless, a methodology specially designed to train the distribution for anomaly detection has not been suggested yet. To address issues, we propose a proper method to update the base distribution for anomaly detection tasks.

\section{Preliminary: Normalizing flow}
\subsection{Notation}
$\theta$ and $\psi$ represent the parameters of the normalizing flow and that of the base distribution, respectively. We use $p(\cdot)$ as a distribution and the probability density function of the distribution. $p_{Z}(\cdot; \psi)$ indicates the base distribution and $p_X(\cdot; \theta, \psi)$ denotes the distribution induced from the base distribution by normalizing flow. In addition, $p_X^*(\cdot)$ is the data distribution that the model aims to learn, e.g. distribution of images.

\subsection{Fitting methodology}
The idea of normalizing flow is from a change of random variables as
\begin{equation}
\label{eq-change-of-variable}
    p_{X}({x}) = p_Z({z}) \left|\det \frac{\partial {z}}{\partial {x}}\right|.
\end{equation}
This implies that a complex distribution $p_X(\cdot; \theta, \psi)$ can be factorized to simple and known distribution $p_Z(\cdot; \psi)$ using invertible mapping $f_\theta :X \rightarrow Z$. In general, the normalizing flow stack the invertible neural layers, $f_\theta = f_{\theta_1} \circ f_{\theta_2} \circ \cdots \circ f_{\theta_n}$ to form the invertible mapping, where $f_{\theta_i}$ is the layer of neural network.

To find the correct invertible mapping $f_\theta$ with respect to the $Z$ and $X$, we can use maximum likelihood estimation: estimating the parameters of distribution with random samples from the distribution. In particular, maximizing the likelihood can be done by maximizing log-likelihood, which is
\begin{equation}\label{eq:log-likelihood}
    \log{p_{X}(x; \theta, \psi)}
    =\log{p_{Z}(f_\theta(x); \psi)}+\log\left| \det\left( \frac{\partial{f_\theta(x)}}{\partial{x}} \right) \right|.
\end{equation}
This objective function is based on the fact that maximizing the log-likelihood is equivalent to minimizing Monte Carlo approximation of $\kl{p_X ^*(x)}{p_X(x;\theta, \psi)}$ on sample $x$ from $p_X^*(x)$ \cite{papamakarios1}.

\section{AltUB}
In this section, we introduce AltUB, an effective method to update the base distribution of normalizing flow to enhance the expressive power of the flow and increase stability in detecting anomalies. We first state the limitations of prior normalizing flow-based works for anomaly detection, which affect the stability. Then, we propose our method.

\subsection{Limitations of Non-trainable base distribution of normalizing flow}
\label{sec4.1}
Normalizing flow-based anomaly detection models \cite{rudolph1, yu1, gudovskiy1} define normalizing flow $f_\theta : X \rightarrow Z$ as a mapping from extracted features of images to simple, fixed distribution. e.g. $N(0, I)$. The normalizing flow architecture is used for learning a distribution of representation of normal samples to detect out-of-distribution samples. However, Figure \ref{figure-fastflow-dist} shows that the models seem to transform normal images into not $N(0, I)$ as their purpose, but into several different normal distributions of $CHW$ dimensions due to their low expressive power. 

The most simple way to enhance the expressive power is to stack more layers. In fact, Komologorov-Smirnov statistics and the train loss(negative log-likelihood) decrease as flow gets deeper as shown in Figure \ref{figure-fastflow-depth} (a) and Figure \ref{figure-fastflow-depth} (b). Nevertheless, the average of mean square tends to increase while adding more layers for shallow depth, which indicates the phenomenon of Figure \ref{figure-fastflow-dist} (mean-shift) occurs. We guess the reason as follows: while an arbitrary distribution is transformed toward $N(0, 1)$ along multiple layers of flow, the conversion into the normal distribution takes precedence over the proper adjustment of specific parameters as mean and variance.

Figure \ref{figure-fastflow-depth} (a) represents that the mean-shift decreases in the deeper depth than 8. We are able to easily figure out that the mean-shift phenomenon can be solved by increasing expressive power. However, stacking layers may not lead to higher performance (see Figure \ref{figure-fastflow-depth} (c)) because of overfitting. In the other words, stacking more and more layers may disrupt learning the distribution of normal samples, even if it increases expressive power. Therefore, the flow model with non-trainable base distribution cannot break through the mean-shift simply.

\subsection{Trainable base distribution of normalizing flow}
As Figure \ref{figure-fastflow-dist} and Section \ref{sec4.1} show, a method that increases expressive power with a few layers is required to match the model output and score function. This method may improve the performance by using the right score function to find anomalies.

Meanwhile, \cite{mask-auto} proves that the following equation holds:
\begin{equation}
    \label{eq-kl-base}
    \kl{p_X ^* ({x})}{p_{X}({x; \theta, \psi})} = \kl{p_{Z}^*({z; \theta})}{p_{Z}({z; \psi})},
\end{equation}
where $p_X ^* ({\cdot})$ is a target distribution in $X$ and $p_Z^* (\cdot; \theta)$ is the induced from $p_X ^* ({\cdot})$ by normalizing flow $f_\theta$. The equation \ref{eq-kl-base} means that the effective update of $p_{Z}({z; \psi})$ toward $p_{Z}^*({z; \phi})$ leads $p_{X}({x; \theta})$ to be similar as $p_X ^* ({x})$. Therefore, we can increase the expressive power of models by updating the base distribution. Based on this fact, AltUB assumes the base distribution of the models to be $N(\mu, \Sigma)$, i.e. normal distribution with learnable parameters of CHW dimensions.

In our method, along with the parameters of normalizing flow $\theta$, the parameters of base distribution $\psi$ are also trained with stochastic gradient-based methods using the gradient of negative log-likelihood (loss) as follows: \small
\begin{equation}
\begin{split}
    \label{eq:grad_log-likelihood}
    &\nabla_\theta L(\theta, \psi) = -\nabla_\theta\log p_Z(f_\theta(x);\psi) - \nabla_\theta\log\left| \det\left( \frac{\partial{f_\theta(x)}}{\partial{x}} \right) \right| \\
    &\nabla_\psi L(\theta, \psi) = -\nabla_\psi\log p_Z(f_\theta(x);\psi),
\end{split}
\end{equation}
\normalsize
where $L(\theta, \psi) = -\log p_X(x;\theta, \psi)$ is the negative log-likelihood of samples from $X$.

As an anomaly score of a test sample with $C$ channels, we use the likelihoods in the base distribution: \small
\begin{equation}
    -\exp\left({\frac{-1}{2|C|}}\sum_{c \in C}{ \left({(z_c-\mu_c)^T\Sigma^{-1}(z_c-\mu_c)} + \ln |\det\Sigma_c| \right)}\right),
\end{equation}
\normalsize
Because we obtain the appropriate parameters of the base distribution from AltUB, our score function will detect the anomalies more precisely.

\subsection{The method to train base distribution}
When we assume the base distribution as $N(\mu, \Sigma)$, the gradient of negative log-likelihood of Equation \ref{eq:grad_log-likelihood} becomes as follows: \small
\begin{equation}
\begin{split}
    &\nabla_\mu L(\theta, \mu, \Sigma) = \Sigma^{-1} (\mu-f_\theta(x)) \\
    &\nabla_\Sigma L(\theta, \mu, \Sigma) = {\frac{1}{2}}\Sigma^{-1}\left(I - (f_\theta(x)-\mu) (f_\theta(x)-\mu)^T \Sigma^{-1} \right).
\end{split}
\end{equation}
\normalsize
The loss function consists of subtraction, hence the gradients will be small. However, we observe that some outputs like Figure \ref{figure-fastflow-dist} (c) are quite far from $N(0, 1)$. Therefore, the parameters of the base distribution are not able to be trained effectively with the original learning rate of the main models, e.g. 0.001 of FastFlow. In addition, because the normalizing flow model learns slowly to find the distribution of representations of normal images, 
a consistently high learning rate for parameters of base distribution will harm the model.

To train the base distribution well yet not harm the model to learn, the alternating training method is introduced. In general, a normalizing flow model and the base distribution are trained together with the original optimizer and learning rate $\eta_1$. But only the base distribution is trained with the larger learning rate $\eta_2$ using an SGD optimizer once in a freezing interval. The detail is depicted in Algorithm \ref{algorithm-altub}.

Due to the high dimensionality of the space for base distribution, AltUB may suffer from computational complexity when using $\Sigma$. This is because of the computational complexity to calculate $\Sigma^{-1}$ in the gradients (about $O(n^{3})$). Therefore, we assume the independence of each dimension of the base distribution, as \cite{rudolph1, yu1, gudovskiy1} did. i.e. $\Sigma$ is assumed to be the diagonal matrix.

AltUB trains the value of parameters of the base distribution effectively, hence increasing the expressive power while helping to detect anomalies. Because AltUB is simple and independent of an upstream model, the proposed method can be applied to any normalizing flow model for anomaly detection. 
\begin{algorithm}[t]
   \caption{AD flow+AltUB}
   \label{algorithm-altub}
\begin{algorithmic}
    \STATE {\bfseries Input:} Pre-trained feature extractor $\phi$, normal data $\mathcal{X}$, epoch $E$
    \STATE{$\theta$: Parameters of an AD Flow}
    \STATE{$\psi$: $\left\{\mu, \ln{\Sigma}^{\frac{1}{2}}\right\}$} \COMMENT{Initialized to zero.}
    \FOR{batch B $in$ $\mathcal{X}$} 
        \STATE {Feature $\leftarrow \phi$(B)}
        \STATE {Output $\leftarrow f_\theta$(Feature)}
        \STATE {Loss $\leftarrow$ $-\ln(Likelihood(\textrm{Output}, N(\mu, \Sigma))$}
        \newline
        \IF{$\textrm{E} \not\equiv 0 \: \mathrm{mod} \: \textrm{FreezingInterval} $}
            \STATE $Update(\left\{\theta, \psi\right\}, \textrm{lr}: \eta_1)$
        \ELSE
            \STATE {$\psi \leftarrow \psi - \eta_2 \nabla\psi$}
            \ENDIF
    \ENDFOR
\end{algorithmic}
\end{algorithm}

\section{Experiments}

\subsection{Experimental Setups}
\begin{table*}[t] 
\caption{Official anomaly detection and segmentation performance on MVTec AD dataset with the format (detection AUROC, segmentation AUROC). The performance of the proposed method is recorded as the best one during the training process. For FastFlow, the best one is chosen between the four distinct feature extractors. The performance with bold letters means the best one among these models.}
\label{table-altub-best}
\centering
\begin{tabular}{c|c|c|c|c|c|c|c|cc}
\hline
Model      & SSPCAB      & PatchCore   & CS-Flow  & DifferNet & CFLOW-AD      & FastFlow     & \multicolumn{2}{c}{Flow + AltUB (ours)}                                  \\ \hline
Base model & DRAEM       & -           & -        & -         & -           & -            & \multicolumn{1}{c|}{CFLOW-AD}       & \multicolumn{1}{c}{FastFlow} \\ \hline
Carpet     & (98.2,95.5) & (98.7,99.1) & (99.0,-) & (92.9,-)  & (\textbf{99.3},99.3) & (100,99.4)         & \multicolumn{1}{c|}{(99.2,99.3)} & \multicolumn{1}{c}{(-,\textbf{99.5})} \\
Grid       & (\textbf{100,99.7})  & (98.6,98.7) & (\textbf{100},-)  & (84.0,-)  & (99.6,99.0) & (99.7,98.3)          & \multicolumn{1}{c|}{(\textbf{100},99.1)}  & \multicolumn{1}{c}{(-,99.3)} \\
Leather    & (\textbf{100},99.5)  & (\textbf{100},99.3)  & (\textbf{100},-)  & (97.1,-)  & (\textbf{100,99.7})  & (\textbf{100},99.5)       & \multicolumn{1}{c|}{(\textbf{100}, \textbf{99.7})}  & \multicolumn{1}{c}{(-,\textbf{99.7})} \\
Tile       & (\textbf{100,99.3})  & (99.4,96.1) & (\textbf{100},-)  & (99.4,-)  & (99.9,98.0) & (\textbf{100},96.3)          & \multicolumn{1}{c|}{(99.9,98.0)} & \multicolumn{1}{c}{(-,97.6)} \\
Wood       & (99.5,96.8) & (99.2,95.1) & (\textbf{100},-)  & (99.8,-)  & (99.1,96.7) & (\textbf{100},97.0)      & \multicolumn{1}{c|}{(99.0,96.6)} & \multicolumn{1}{c}{(-,96.9)} \\
Bottle     & (98.4,98.8) & (\textbf{100},98.6)  & (99.8,-) & (99.0,-)  & (\textbf{100, 99.0}) & (\textbf{100},97.7))    & \multicolumn{1}{c|}{(\textbf{100},99.0)}  & \multicolumn{1}{c}{(-,\textbf{99.0})} \\
Cable      & (96.9,96.0) & (99.5,\textbf{98.5}) & (99.1,-) & (95.9,-)  & (97.6,97.6) & (\textbf{100},98.4)        & \multicolumn{1}{c|}{(97.8,97.6)} & \multicolumn{1}{c}{(-,98.4)} \\
Capsule    & (99.3,93.1) & (98.1,98.9) & (97.1,-) & (86.9,-)  & (97.7,99.0) & (\textbf{100,99.1})       & \multicolumn{1}{c|}{(98.1,99.0)} & \multicolumn{1}{c}{(-,\textbf{99.1})} \\
Hazelnut   & (\textbf{100.99.8})  & (\textbf{100},98.7)  & (99.6,-) & (99.3,-)  & (\textbf{100},98.9)  & (\textbf{100},99.1)         & \multicolumn{1}{c|}{(\textbf{100},98.9)}  & \multicolumn{1}{c}{(-,99.3)} \\
Metalnut  & (\textbf{100,98.9})  & (\textbf{100},98.4)  & (99.1,-) & (96.1,-)  & (99.3,98.6) & (\textbf{100},98.5)       & \multicolumn{1}{c|}{(99.5,98.6)} & \multicolumn{1}{c}{(-,98.7)} \\
Pill       & (\textbf{99.8},97.5) & (97.0,97.6) & (98.6,-) & (88.8,-)  & (96.8,99.0) & (99.4,\textbf{99.2})         & \multicolumn{1}{c|}{(97.0,99.0)} & \multicolumn{1}{c}{(-,99.1)} \\
Screw      & (97.9,\textbf{99.8}) & (\textbf{98.1},99.4) & (97.6,-) & (96.3,-)  & (91.9,98.9) & (97.8,99.4)   & \multicolumn{1}{c|}{(91.7,98.9)} & \multicolumn{1}{c}{(-,99.5)}        \\
ToothBrush & (\textbf{100},98.1)  & (\textbf{100},98.7)  & (91.9,-) & (98.6,-)  & (99.7,98.9) & (94.4,98.9)     & \multicolumn{1}{c|}{(99.4,98.9)} & \multicolumn{1}{c}{(-,\textbf{99.2})} \\
Transistor & (92.9,87.0) & (\textbf{100},96.4)  & (99.3,-) & (91.1,-)  & (95.2,98.0) & (99.8,97.3)   & \multicolumn{1}{c|}{(95.2,\textbf{98.2})} & \multicolumn{1}{c}{(-,98.0)} \\
Zipper     & (\textbf{100},99.0)  & (99.5,98.9) & (99.7,-) & (95.1,-)  & (98.5,\textbf{99.1}) & (99.5,98.7)      & \multicolumn{1}{c|}{(98.5,99.1)} & \multicolumn{1}{c}{(-,\textbf{99.1})} \\ \hline
Overall      & (98.9, 97.1) & (99.1,98.1) & (98.7,-) & (94.9,-)  & (98.3,98.6) & (\textbf{99.4},98.5)      & \multicolumn{1}{c|}{(98.4,98.7)} &  \multicolumn{1}{c}{(-,\textbf{98.8})} \\ \hline
\end{tabular}
\end{table*}

\subsubsection{Datasets}
We evaluate our method on two datasets: MVTec AD \cite{bergmann1} and beanTech Anomaly Detection(BTAD) \cite{btad} which are well-known as benchmark of anomaly detection models.

MVTec AD is a widely utilized dataset specifically designed for unsupervised anomaly detection. It consists of 5354 high-resolution images of industrial objects or patterns. In detail, it is composed of a total of 15 categories with 3629 images for training and validation and 1725 images for testing. We compare the proposed method with SSPCAB+DRAEM \cite{sspcab, draem}, PatchCore \cite{patchcore}, and CS-Flow \cite{cs-flow} which are current state-of-the-art models. 

BTAD is a real-world dataset for industrial unsupervised anomaly detection. It comprises 2540 images divided into three categories of industrial products. The performance of our method on BTAD is compared with PatchSVDD \cite{patchsvdd}, VT-ADL \cite{vt-adl}, and the method introduced in \cite{mspbrl}.

Both MVTec AD and BTAD have only normal images in the training set whereas normal images and abnormal images are together in the test set.

\subsubsection{Metrics}
The performance of models is measured by the Area Under the Receiver Operating Characteristic curve (AUROC) at the image or pixel level. which is a general metric for the evaluation of anomaly detection models. Meanwhile, the expected value of the performance is more practical than the best performance during training in unsupervised tasks because the validation set is not provided in the real world. In particular, the variance of the performance is also significant for stability. To achieve the purpose of real-world anomaly detection tasks, both the best and the average value of AUROC evaluated during the training process (from 100th epochs to 300th epochs) are adopted for comparison of normalizing flow-based models. We provide (mean, standard deviation) for average performance.

\subsection{Anomaly detection and segmentation performance} 
\begin{table*}[t]
\caption{Anomaly detection and segmentation performance on MVTec AD dataset with the format  (average of AUROC, standard deviation of AUROC). AUROCs are measured from the 100th epoch. For FastFlow, performances with the same feature extractor in Table \ref{table-altub-best} are recorded for each category. The better performance between an original model and the one with AltUB is marked in bold letters.}
\label{table-altub-stability}
\centering
\begin{tabular}{c|cccc|cc}
\hline
Task type  & \multicolumn{4}{c|}{Segmentation}                                                                                                                                    & \multicolumn{2}{c}{Detection}                           \\ \hline
Base model & \multicolumn{2}{c|}{FastFlow}                                                         & \multicolumn{4}{c}{CFLOW-AD}                                                                                                           \\ \hline
           & \multicolumn{1}{c}{-}                     & \multicolumn{1}{c|}{+ AltUB}               & \multicolumn{1}{c}{-}            & \multicolumn{1}{c|}{+ AltUB}               & \multicolumn{1}{c}{-}             & + AltUB               \\ \hline
Carpet     & \multicolumn{1}{c}{(98.5,0.22)}          & \multicolumn{1}{c|}{\textbf{(98.8,0.13)}} & \multicolumn{1}{c}{(99.2,0.09)} & \multicolumn{1}{c|}{(99.2,0.08)}          & \multicolumn{1}{c}{(97.6,0.24)}  & (97.6,0.30)          \\
Grid       & \multicolumn{1}{c}{(99.2,0.17)}          & \multicolumn{1}{c|}{\textbf{(99.3,0.02)}} & \multicolumn{1}{c}{(98.7,0.08)} & \multicolumn{1}{c|}{(98.7,0.08)}          & \multicolumn{1}{c}{(95.5,1.77)}  & \textbf{(96.0,1.51)} \\
Leather    & \multicolumn{1}{c}{(99.6,0.03)}          & \multicolumn{1}{c|}{\textbf{(99.7,0.01)}} & \multicolumn{1}{c}{(99.6,0.04)} & \multicolumn{1}{c|}{(99.6,0.04)}          & \multicolumn{1}{c}{(100,0.02)}   & (100,0.03)           \\
Tile       & \multicolumn{1}{c}{(96.9,0.55)}          & \multicolumn{1}{c|}{\textbf{(97.0,0.44)}} & \multicolumn{1}{c}{(97.8,0.17)} & \multicolumn{1}{c|}{(97.8,0.13)}          & \multicolumn{1}{c}{(99.2,0.22)}  & \textbf{(99.3,0.17)} \\
Wood       & \multicolumn{1}{c}{(95.4,0.57)}          & \multicolumn{1}{c|}{\textbf{(96.4,0.21)}} & \multicolumn{1}{c}{(96.1,0.56)} & \multicolumn{1}{c|}{\textbf{(96.2,0.50)}} & \multicolumn{1}{c}{\textbf{(98.6,0.16)}}  & (98.3,0.19)          \\
Bottle     & \multicolumn{1}{c}{(98.6,0.56)}          & \multicolumn{1}{c|}{\textbf{(98.9,0.06)}} & \multicolumn{1}{c}{(98.9,0.05)} & \multicolumn{1}{c|}{(98.9,0.06)}          & \multicolumn{1}{c}{(100,0.00)}   & (100,0.00)           \\
Cable      & \multicolumn{1}{c}{(97.9,0.11)}          & \multicolumn{1}{c|}{\textbf{(98.1,0.08)}} & \multicolumn{1}{c}{(97.5,0.05)} & \multicolumn{1}{c|}{(97.5,0.08)}          & \multicolumn{1}{c}{(96,3,0.42)}  & \textbf{(97.2,0.22)} \\
Capsule    & \multicolumn{1}{c}{(98.7,0.07)} & 
\multicolumn{1}{c|}{\textbf{(98.9,0.05)}}          & \multicolumn{1}{c}{\textbf{(99.0,0.03)}} & \multicolumn{1}{c|}{(98.9,0.04)}          & \multicolumn{1}{c}{(97.2,0.53)}  & \textbf{(97.5,0.34)} \\
Hazelnut   & \multicolumn{1}{c}{(98.8,0.17)}          & \multicolumn{1}{c|}{\textbf{(98.9,0.13)}} & \multicolumn{1}{c}{(98.8,0.03)} & \multicolumn{1}{c|}{(98.8,0.02)}          & \multicolumn{1}{c}{(100,0.00)}   & (100,0.00)           \\
Metal nut  & \multicolumn{1}{c}{(97.8,0.18)}          & \multicolumn{1}{c|}{\textbf{(97.9,0.19)}} & \multicolumn{1}{c}{(98.5,0.06)} & \multicolumn{1}{c|}{(98.5,0.04)}          & \multicolumn{1}{c}{(98.4,0.30)}  & \textbf{(98.7,0.34)} \\
Pill       & \multicolumn{1}{c}{(97.9,0.28)}          & \multicolumn{1}{c|}{\textbf{(98.8,0.10)}} & \multicolumn{1}{c}{(98.9,0.05)} & \multicolumn{1}{c|}{(98.9,0.02)}          & \multicolumn{1}{c}{(94.4,1.20)}  & \textbf{(94.7,0.76)} \\
Screw      & \multicolumn{1}{c}{(99.3,0.15)}          & \multicolumn{1}{c|}{\textbf{(99.4,0.11)}} & \multicolumn{1}{c}{(98.7,0.05)} & \multicolumn{1}{c|}{(98.7,0.07)}          & \multicolumn{1}{c}{(85.2,1.76)}  & \textbf{(88.8,1.29)} \\
ToothBrush & \multicolumn{1}{c}{(98.9,0.09)}          & \multicolumn{1}{c|}{\textbf{(99.0,0.06)}} & \multicolumn{1}{c}{(98.9,0.29)} & \multicolumn{1}{c|}{(98.9,0.08)}          & \multicolumn{1}{c}{(94.1,1.28)}  & \textbf{(95.6,1.16)}          \\
Transistor & \multicolumn{1}{c}{(96.2,1.32)}          & \multicolumn{1}{c|}{\textbf{(96.8,0.21)}} & \multicolumn{1}{c}{(98.0,0.09)} & \multicolumn{1}{c|}{\textbf{(98.1,0.07)}} & \multicolumn{1}{c}{\textbf{(94.4,0.46)}}  & (93.3,0.79)          \\
Zipper     & \multicolumn{1}{c}{(98.6,0.29)}          & \multicolumn{1}{c|}{\textbf{(98.7,0.12)}} & \multicolumn{1}{c}{(99.0,0.07)} & \multicolumn{1}{c|}{(99.0,0.07)}          & \multicolumn{1}{c}{(96.7,0.23)}  & (96.7,0.26)          \\ \hline
Overall    & \multicolumn{1}{c}{(98.2,0.33)}          & \multicolumn{1}{c|}{\textbf{(98.4,0.13)}} & \multicolumn{1}{c}{(98.5,0.11)} & \multicolumn{1}{c|}{(98.5,0.09)}          & \multicolumn{1}{c}{(96.5, 0.57)} & \textbf{(96.9,0.49)} \\ \hline
\end{tabular}
\end{table*}

\begin{table}[t]
\caption{Official anomaly segmentation performance on BTAD dataset. The best performances during the training process of the proposed method are recorded for each category. WRN50 is utilized as the feature extractor.}
\label{table-btad-best}
\centering
\small{
\begin{tabular}{c|c|c|c|cc}
\hline
Model & P-SVDD & VT-ADL & Tsai et al.   & \multicolumn{2}{c}{FastFlow}          \\ \hline
      & -      & -      & -             & \multicolumn{1}{c}{-} & +AltUB        \\ \hline
1     & 94.9   & 76.3   & \textbf{97.3} & 95                     & 97.1          \\
2     & 92.7   & 88.9   & 96.8          & 96                     & \textbf{97.6}          \\
3     & 91.7   & 80.3   & 99.0          & 99                     & \textbf{99.8} \\ \hline
Overall  & 93.1   & 81.8   & 97.7          & 97                     & \textbf{98.2} \\ \hline
\end{tabular}
}
\end{table}

\begin{table}[t]
\caption{Anomaly segmentation performance on BTAD dataset with the format  (average of AUROC, standard deviation of AUROC). AUROCs are measured from the 100th epoch. As Table \ref{table-btad-best}, WRN50 is utilized as the feature extractor. The better performance between an original model and the one with AltUB is marked in bold letters.}
\label{table-btad-stability}
\centering
\begin{tabular}{c|cc}
\hline
AltUB                    & w/o                              & w                                         \\ \hline
1                        & (94.3, 0.71)                     & \textbf{(96.7, 0.12)}                     \\
2                        & (95.8, 0.23)                     & \textbf{(96.2, 0.19)}                     \\
3                        & \textbf{(99.2, 0.11)}            & (98.9, 0.16)                              \\ \hline
\multicolumn{1}{c|}{Overall} & \multicolumn{1}{l}{(96.4, 0.35)} & \multicolumn{1}{l}{\textbf{(97.3, 0.16)}} \\ \hline
\end{tabular}
\end{table}

\begin{figure}[t]
\begin{center}
\centerline{\includegraphics[width=\columnwidth]{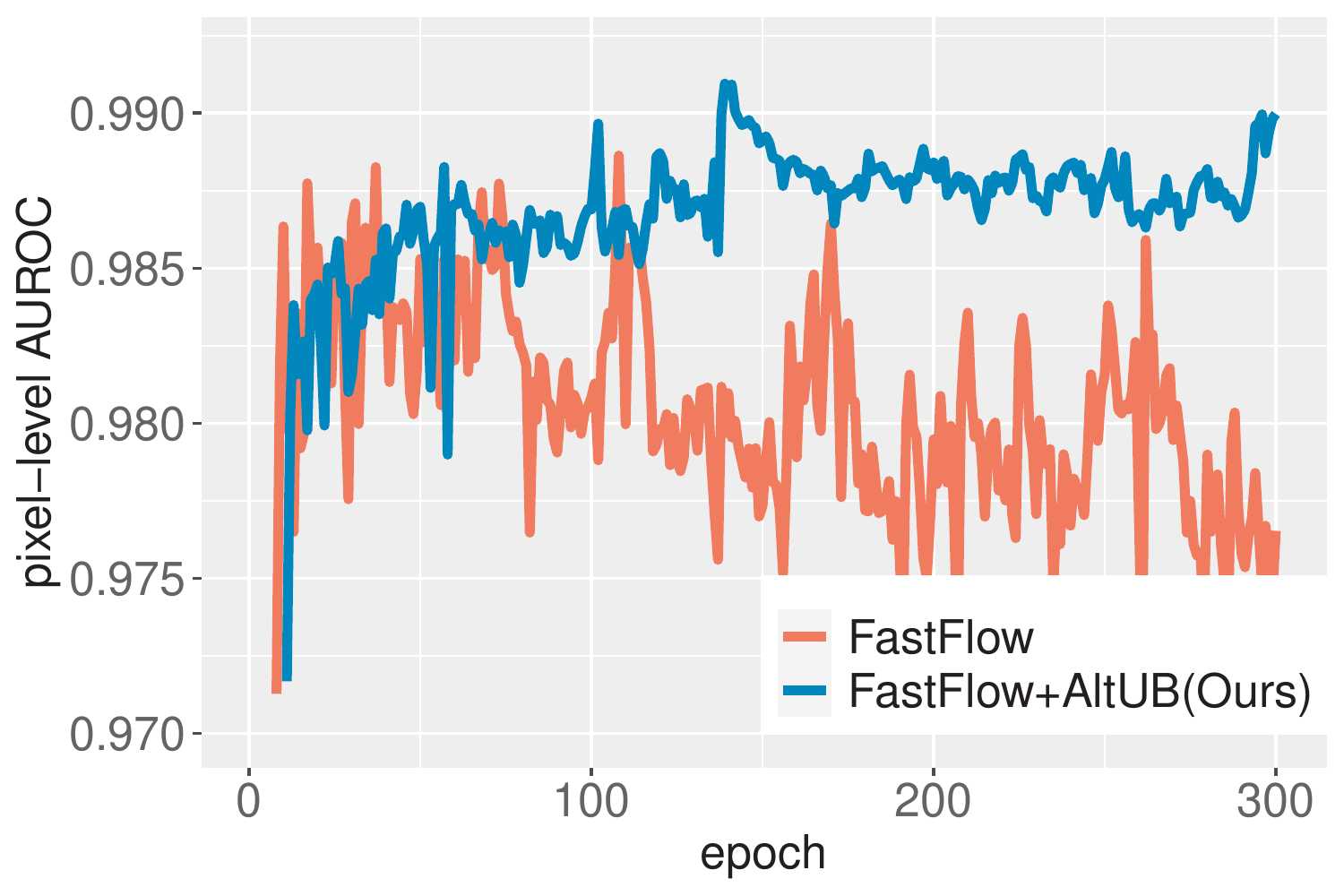}}
\caption{The pixel-wise AUROC is measured on the pill category of the MVTec AD dataset. FastFlow based on CaiT and that applied AltUB are compared.}
\label{figure-fastflow-auroc}
\end{center}
\end{figure}

We apply AltUB to FastFlow and CFLOW-AD which are well-known and well-performed NF models for anomaly detection. To be specific, we compare the performance of these original models and those of applied AltUB. ResNet18 (R18) \cite{resnet}, WideResNet50 (WRN50) \cite{wideresnet}, CaiT \cite{cait}, and DeiT \cite{deit} for FastFlow, WRN50 for CFLOW-AD are utilized for feature extractors. We use anomalib \cite{anomalib} for FastFlow implementation because the official code of FastFlow is not provided yet. We do not perform detection with FastFlow as the way how to define anomaly score at image level is not clear.

\subsubsection{MVTec AD}
First of all, the best performance during the training process is compared in Table \ref{table-altub-best}. Our method with FastFlow clearly achieves the new state-of-the-art performance on the anomaly segmentation task of the MVTec AD dataset. In addition, AltUB improves the stability of performance of NF models as Table \ref{table-altub-stability}, in terms of the point that AltUB enhances the average performance and reduces its variance in most cases. One sample is depicted in Figure \ref{figure-fastflow-auroc}.
This observation is consistent with FastFlow based on the other feature extractors. The full information about the performance is provided in the supplementary material \ref{table-fastfow-full}.
\subsubsection{BTAD}
For generalization, we evaluate the proposed method with BTAD. As Table \ref{table-btad-best} and Table \ref{table-btad-stability}, AltUB also increases both the best performance and the stability of performance on BTAD except in one case. Our method also establishes the state-of-the-art performance on the anomaly segmentation task of BTAD.

\subsection{Examination for normalization}
To examine whether AltUB is effective to update the base distribution, a one-sample Kolmogorov–Smirnov test (KS-test) is utilized. One-sample KS statistic for an empirical cumulative density function (cdf) $F_n$ and a given cdf $F$ is defined as 
\begin{equation}\label{eq:Two-sample_KS_statistic}
KS(F_{n}, F) = \sup_x\left|F_{n}(x) - F(x)\right| \in [0, 1],
\end{equation}
where $n$ is the number of independent and identically distributed samples for the empirical distribution.
$KS(\cdot, \cdot)$ converges to 0 as the two distribution is similar. Ideally, each spatial location of each channel of outputs corresponding to normal inputs shows the $N(0, 1)$.

Let $z=f_\theta(\phi(B))$ be the output of flow model with the shape $[B, C, H, W]$. Then, we can expect that $z[:,c,h,w] \sim N(0, 1)$ for every $c, h, w$. This also implies that $z[:,c,:,:]$ should follow $N(0, 1)$.

For an original flow model, $N(0, 1)$ and $z[:,c,,]$ are compared while $N(0, 1)$ and $\left(z[:,c,:,:]-\mu[c,:,:]\right)/\sigma[c,:,:]$ are compared for our method by the KS-statistics. Table \ref{table-fastflow-kstest} apparently shows that our method achieves the lower KS statistic, meaning that the base distribution is trained effectively. This observation supports our hypothesis that targets the expressive power and reveals the reason why AltUB can achieve higher and more stable performance effectively.

\begin{table}[t]
\caption{95\% confidence interval of Kolmogorov-Smirnov statistics (\%). KS test is applied to each channel of outputs of FastFlow, i.e output[:,C,:,:] once in 10 epochs.}
\label{table-fastflow-kstest}
\centering
\small{
\begin{tabular}{c|c|c|c|c}
\hline
AltUB & R18                 & WRN50            & DeiT                   & CaiT                   \\
\hline
w/o    & 7.39±0.01          & 8.08±0.01          & 7.72±0.04          & 9.13±0.03          \\
w      & \textbf{5.14±0.01} & \textbf{5.65±0.01} & \textbf{3.68±0.01} & \textbf{3.65±0.01} \\
\hline
\end{tabular}
}
\end{table}

\subsection{Comparison with other methods to update the base distribution}
\begin{table}[t]
\caption{Anomaly segmentation performance on the capsule category of MVTec AD dataset with the format (average of AUROC, standard deviation of AUROC). Various methods to update the base distribution are applied to FastFlow. Ster: Stereotype, CNN: A single layer of convolution neural network.}
\label{table-comparison}
\centering
\small{
\begin{tabular}{c|c|c|c|c}
\hline
           & R18         & WRN50       & DeiT        & CaiT        \\ \hline
None       & (98.3,0.09) & \textbf{(98.8,0.09)} & (98.1,0.12)          & (98.7,0.07) \\ 
Ster       & (98.5,0.04) & (98.8,0.10)          & (98.1,0.15)          & (98.6,0.09) \\ 
CNN        & (98.0,0.89) & Crashes              & (97.7,2.40)          & (95.3,7.22) \\ \hline
Ours      &  \textbf{(98.6,0.02)} & (98.6,0.10) & \textbf{(98.6,0.04)} & \textbf{(98.9,0.05)} \\ \hline
\end{tabular}
}
\end{table}

To justify the necessity of alternating training, we compare AltUB with other methods to train the base distribution in Table \ref{table-comparison}. As we mentioned before, the general training (stereotype) with an original learning rate isn't effective as the learning rate isn't large enough. Update using CNN decreases the performance because CNN can't distinguish anomalies and the parameters of the base distribution are sensitive to the weight of CNN. 

\subsection{Inference time} 
\begin{table}[t]
\caption{Inference time (millisecond) per one image comparison between FastFlow with AltUB and without it for the various feature extractors.}
\label{table-fastflow-time}
\centering
\begin{tabular}{c|c|c|c|c}
\hline
AltUB & R18                 & WRN50           & DeiT                   & CaiT                   \\
\hline
w/o    & 2.58 & 12.39 & 15.26 & 234.79          \\
w      & 2.65 & 12.77 & 15.33 & 234.91 \\
\hline
\end{tabular}
\end{table}

Because AltUB doesn't require complex calculations, the inference time has hardly increased. The analysis results are shown in Table \ref{table-fastflow-time}. The inference time is measured by Intel(R) CPU i7-12700K and NVIDIA GeForce RTX 3060.

\subsection{Implementation details}
In our method, $\ln \Sigma ^ {\frac{1}{2}}$ is trained instead of $\Sigma$ to have negative values for itself. In addition, $\mu$ and $\ln \Sigma ^ {\frac{1}{2}}$ are initialized to zero not to disrupt the initial training of the flow model.

We set the freezing interval as 5 epochs and $\eta_2$ as a product of a constant and the original learning rate $\eta_1$, The constant is determined that the maximum value of $\eta_2$ equals 0.05. Exceptionally AltUB is not applied during the warm-up process of CFLOW-AD. We apply gradient clipping as much as 100 to the base distribution for FastFlow to prevent divergence of loss. Layer normalization is applied to outputs of FastFlow with WRN50 and CaiT. We fix the other hyperparameters of flow models as their original papers.

\section{Conclusion}
In this paper, we propose a simple yet effective method AltUB to enhance the expressive power of normalizing flow for anomaly detection. Our key idea is an alternating training method to learn the base distribution. In our experimental results, we prove the proposed method improves the stability of flow models for anomaly detection.

\subsubsection{Limitations and future works}
Despite the success of our method, the performance still fluctuates in some cases. This fact implies that a more suitable training methodology can be suggested. From the point that our simple idea achieves meaningful progress for anomaly detection, an essential approach toward the higher expressive power to detect anomalies will be required.

\bibliography{aaai23}

\appendix
\setcounter{table}{0}
\renewcommand{\thetable}{A\arabic{table}}
\section{Appendix}

We fix a random seed as 25 for the main results of our paper for reproducibility. In addition, We provide more information about performance in Table \ref{table-fastfow-full} (\textbf{Please check the next page}). As our paper, the performance gets higher and more stable in the other feature extractors consistently. 

\begin{table*}[t]

\caption{Anomaly segmentation performance on MVTec AD dataset with the format  (average of AUROC, standard deviation of AUROC). AUROCs are measured from the 100th epochs to 300th epochs. The full performances depending on each feature extractor are provided. On each category, the feature extractor utilized in our paper is marked with an underline.}
\label{table-fastfow-full}
\centering

\begin{tabular}{c|cc|cc|cc|cc}
\hline
Backbone   & \multicolumn{2}{c|}{R18}                    & \multicolumn{2}{c|}{WRN50}                        & \multicolumn{2}{c|}{DeiT}                         & \multicolumn{2}{c}{CaiT}                          \\ \hline
           & -                    & +AltUB               & -                    & +AltUB                     & -                    & +AltUB                     & -                    & +AltUB                     \\ \hline
Carpet     & (98.1,0.20)          & \textbf{(98.6,0.12)} & (98.9,0.34)          & (98.9,0.14)                & {\ul (98.5,0.22)}    & {\ul \textbf{(98.8,0.13)}} & \textbf{(98.8,0.23)} & (98.7,0.10)                \\
Grid       & (98.4,0.11)          & \textbf{(98.6,0.05)} & {\ul (99.2,0.17)}    & {\ul \textbf{(99.3,0.02)}} & (96.6,0.30)          & \textbf{(96.9,0.12)}       & \textbf{(97.0,0.20)}          & (96.9,0.12)                \\
Leather    & (99.5,0.05)          & \textbf{(99.6,0.02)} & {\ul (99.6,0.03)}    & {\ul \textbf{(99.7,0.01)}} & (99.2,0.04)          & \textbf{(99.4,0.03)}       & (99.4, 0.13)         & \textbf{(99.5,0.04)}       \\
Tile       & \textbf{(94.7,0.41)} & (94.1,0.33)          & {\ul (96.9,0.55)}    & {\ul \textbf{(97.0,0.44)}} & (95.7,0.60)          & \textbf{(95.8,0.72)}       & (94.5,0.69)          & \textbf{(94.6,0.29)}                \\
Wood       & (94.0,0.32)          & \textbf{(95.5,0.35)} & {\ul (95.4,0.57)}    & {\ul \textbf{(96.4,0.21)}} & (93.6,0.40)          & \textbf{(95.1,0.25)}       & (94.8,0.43)          & \textbf{(94.9,0.20)}       \\
Bottle     & (96.4,0.13)          & \textbf{(98.6,0.02)} & {\ul (98.6,0.56)}    & {\ul \textbf{(98.9,0.06)}} & (92.1,1.77)          & \textbf{(94.4,0.29)}       & (92.8,0.73)          & \textbf{(96.0,0.55)}       \\
Cable      & (94.6,0.81)          & \textbf{(96.4,0.74)} & (96.9,0.98)          & \textbf{(97.3,0.45)}       & (97.5,0.13)          & \textbf{(97.7,0.08)}       & {\ul (97.9,0.11)}    & {\ul \textbf{(98.1,0.08)}} \\
Capsule    & (98.3,0.09)          & \textbf{(98.6,0.02)} & \textbf{(98.8,0.09)} & (98.6,0.10)                & (98.1,0.12)          & \textbf{(98.6,0.04)}       & {\ul (98.7,0.07)}    & {\ul \textbf{(98.9,0.05)}} \\
Hazelnut   & (95.8,0.57)          & \textbf{(97.2,0.07)} & (96.5,0.93)          & \textbf{(96.8,0.63)}       & \textbf{(98.8,0.13)} & (98.7,0.10)                & {\ul (98.8,0.17)}    & {\ul \textbf{(98.9,0.13)}} \\
Metal nut  & (94.8,0.43)          & \textbf{(97.3,0.06)} & {\ul (97.8,0.18)}    & {\ul \textbf{(97.9,0.19)}} & (97.9,0.17)          & \textbf{(98.4,0.15)}       & (97.4,0.22)          & \textbf{(97.9,0.11)}       \\
Pill       & (95.9,0.16)          & \textbf{(97.3,0.15)} & \textbf{(97.1,0.56)} & (96.9,0.27)                & (98.1,0.28)          & \textbf{(98.4,0.07)}       & {\ul (97.9,0.28)}    & {\ul \textbf{(98.8,0.10)}} \\
Screw      & (94.9,0.84)          & \textbf{(95.7,0.31)} & (97.7,1.01)          & \textbf{(98.7,0.13)}       & (99.0,0.11)          & \textbf{(99.1,0.12)}       & {\ul (99.3,0.15)}    & {\ul \textbf{(99.4,0.11)}} \\
Toothbrush & (96.4,0.22)          & \textbf{(97.8,0.06)} & (97.8,0.12)          & \textbf{(98.3,0.05)}       & (98.0,0.19)          & \textbf{(98.1,0.26)}       & {\ul (98.9,0.09)}    & {\ul \textbf{(99.0,0.35)}} \\
Transistor & (93.0,0.70)          & \textbf{(96.2,0.58)} & {\ul (96.2,1.32)}    & {\ul \textbf{(96.8,0.21)}} & (95.5,0.28)          & \textbf{(96.8,0.17)}       & (95.9,0.30)          & \textbf{(97.4,0.07)}       \\
Zipper     & (98.3,0.16)          & \textbf{(98.5,0.06)} & {\ul (98.6,0.29)}    & {\ul \textbf{(98.7,0.12)}} & (97.2,0.19)          & \textbf{(97.9,0.05)}       & (97.5,0.16)          & \textbf{(98.1,0.07)}       \\ \hline
Overall    & (96.2,0.35)          & \textbf{(97.3,0.20)} & (97.7,0.51)          & \textbf{(98.0,0.20)}       & (97.1,0.33)          & \textbf{(97,6,0.17)}       & (97.3,0.26)          & \textbf{(97.8,0.16)}       \\ \hline
\end{tabular}

\end{table*}

\end{document}